\documentclass[10pt,twocolumn,letterpaper]{article}

\usepackage[pagenumbers]{cvpr} % To force page numbers, e.g. for an arXiv version

\usepackage[dvipsnames]{xcolor}
\usepackage{amsmath}
\usepackage{multirow}

\definecolor{cvprblue}{rgb}{0.21,0.49,0.74}
\usepackage[pagebackref,breaklinks,colorlinks,citecolor=cvprblue]{hyperref}

\def\paperID{*****} % *** Enter the Paper ID here
\def\confName{CVPR}
\def\confYear{2024}

\title{Multi-agent Collaborative Perception for Robotic Fleet: A Systematic Review}

\author{Apoorv Singh\\
Carnegie Mellon University\\
{\tt\small apoorv93singh@gmail.com}
\and
Gaurav Raut\\
University of Maryland\\
{\tt\small gauraut14@gmail.com}
\and
Alka Choudhary\\
Worcester Polytechnic Institute\\
{\tt\small mnnitalka@gmail.com}}

\begin{document}
\maketitle
\begin{abstract}

Collaborative perception in multi-robot fleets is a way to incorporate the power of unity in robotic fleets. Collaborative perception refers to the collective ability of multiple entities or agents to share and integrate their sensory information for a more comprehensive understanding of their environment. In other words, it involves the collaboration and fusion of data from various sensors or sources to enhance perception and decision-making capabilities. By combining data from diverse sources, such as cameras, lidar, radar, or other sensors, the system can create a more accurate and robust representation of the environment. In this review paper, we have summarized findings from 20+ research papers on collaborative perception. Moreover, we discuss testing and evaluation frameworks commonly accepted in academia and industry for autonomous vehicles and autonomous mobile robots. Our experiments with the trivial perception module show an improvement of over 200\% with collaborative perception compared to individual robot perception. Here's our GitHub repository that shows the benefits of collaborative perception: \textbf{https://github.com/synapsemobility/synapseBEV}

% Change Log
% - Added other models (legacy) in abstract
% - et al is used to reference authors. Changed that to etc
\end{abstract}
\section{Introduction}

Collaborative perception in a multi-robot fleet refers to a system where multiple agents or robots work together by sharing perceived data through a communication network to enhance their perception capabilities beyond what a single agent can achieve. This collaborative approach allows robots to perceive their surrounding environment beyond their individual line-of-sight and field-of-view, enabling them to address challenges like long-range scenarios and occlusions that are difficult for a single agent to solve. The concept involves sharing information among vehicles to improve perception performance, especially when individual vehicles have limited perception abilities. 

\begin{figure}[!ht]
  \centering
  \includegraphics[width=0.6\textwidth]{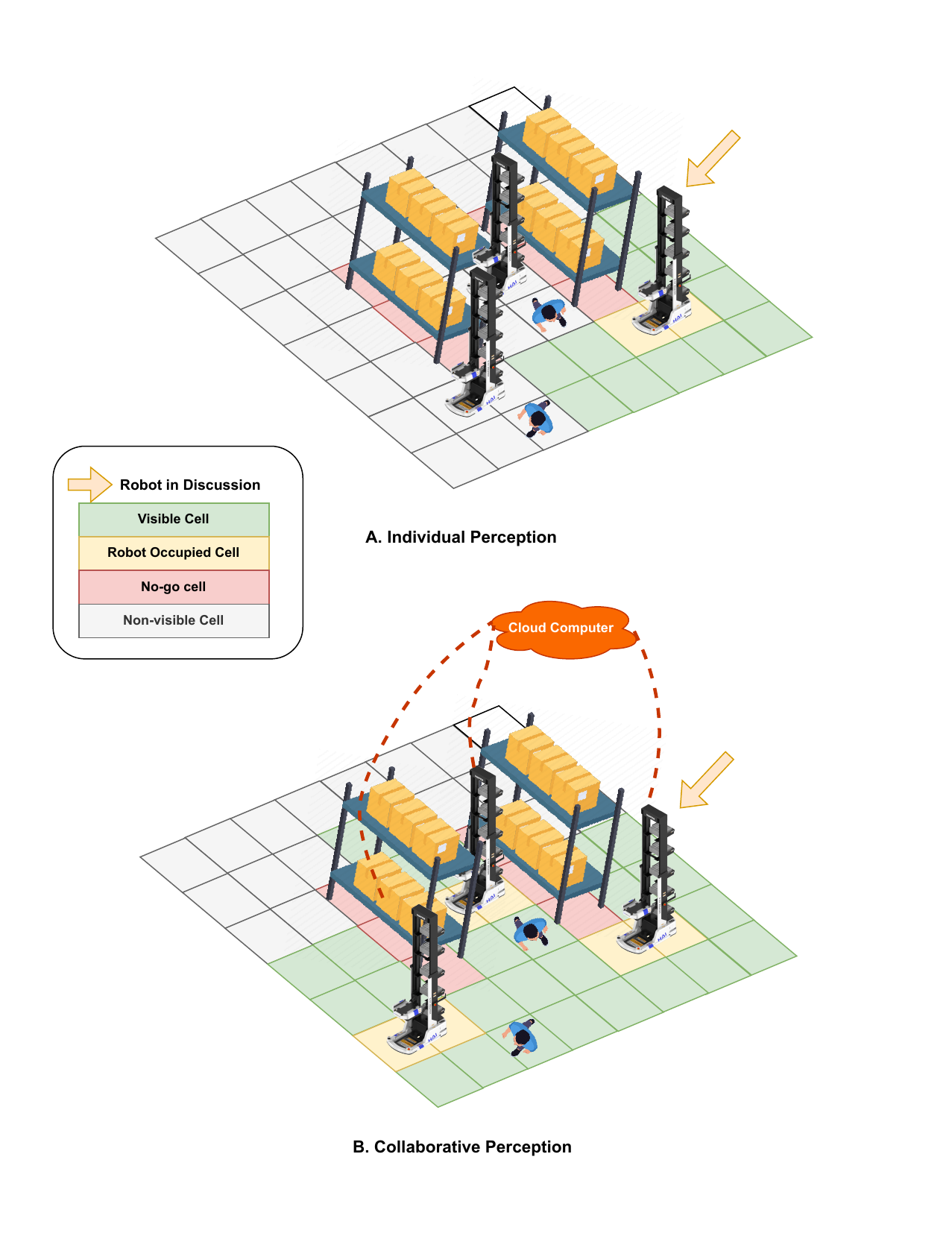}
  \caption{A warehouse use case of collaborative perception. Using collaborative perception, two critical perception challenges, (1) Occlusions and (2) Long-range detections, can be resolved.}
  \label{fig:warehouse_1}
\end{figure}

\begin{figure*}[!ht]
  \centering
  \includegraphics[width=0.75\textwidth]{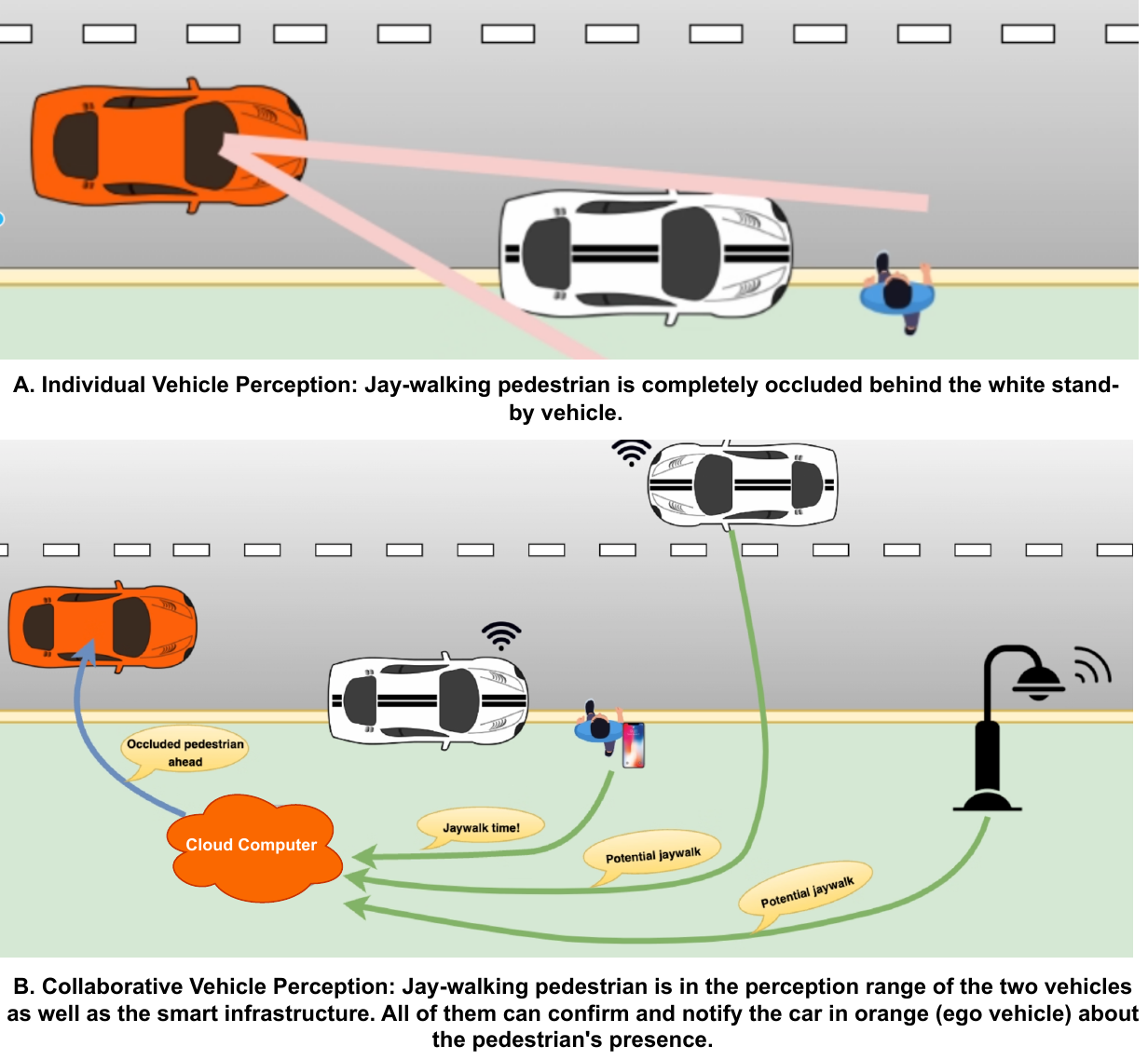}
  \caption{Visualization of collaborative perception. Orange cars (Ego-car) won't have any blind spots once other smart cars in white and smart infrastructure in black share data about the potential collision with the jaywalking pedestrian.}
  \label{fig:car_1}
\end{figure*}

In the context of autonomous driving/ mobile robots, collaborative perception has been proposed as a solution to overcome the limitations of individual vehicles by allowing them to share information and perceive environments collectively. This collaborative approach is crucial for addressing occlusion and sensor failure issues common in autonomous driving systems. By leveraging collaborative perception, vehicles can share information beyond their immediate surroundings, enhancing their ability to detect obstacles, make decisions, and navigate safely in complex environments. In this survey paper, we focus on collaborative perception applications for AMRs (Autonomous mobile robots) working in warehouses, as shown in Fig \cref{fig:warehouse_1} and AVs (Autonomous vehicles) operating on the roads, as shown in Fig. \cref{fig:car_1}. \\

Collaborative perception in multi-robot fleets has various applications that enhance the perception capabilities of individual agents through shared information. Some key applications include:
\begin{itemize}
    \item \textbf{Improved Perception Performance}: Collaborative perception allows robots to share information beyond their individual line-of-sight and field-of-view, enhancing their ability to detect obstacles, make decisions, and navigate safely in complex environments
    \item \textbf{Addressing Occlusion and Sensor Failure}: By sharing collective perception messages (CPM) among vehicles, collaborative perception helps overcome challenges like occlusion and sparse data in long-range scenarios, which are common issues in autonomous driving systems
    \item \textbf{Reduced Cost of Perception Equipment}: Autonomous vehicles are typically equipped with high-fidelity sensors for reliable perception, leading to high costs. Collaborative perception can alleviate this by enabling vehicles to share information with nearby vehicles and infrastructure, reducing the need for expensive individual sensor setups
    \item \textbf{Enhanced Accuracy and Safety}: Studies have shown that collaborative perception among vehicles can improve the accuracy of environmental perception, as well as enhance the robustness and safety of transportation systems
    \item \textbf{Real-Time Collaboration}: Collaborative perception enables vehicles to collaborate in real-time by addressing challenges related to communication capacity and noise, ensuring efficient sharing of information for enhanced perception capabilities
\end{itemize}

\vskip 0.2in
The main contributions of this work can be summarized as follows: 
\begin{itemize}
    \item An overview of collaborative perception for the multi-agent environments to get readers up to speed with the theoretical prerequisites for going through the latest trends in this field. 
    \item In-depth survey on the SOTA approaches for collaborative perception.
    \item Detailed walkthrough on testing and evaluation methods, including simulators and open-source dataset used to evaluate the above SOTA methods.
    \item Highlight current research gaps and future research direction to inspire researchers to advance this new field of multi-agent collaborative perception.
\end{itemize}

\section{Collaborative Perception}
\subsection{Stages of Collaboration}
The 'Collaboration' part of the collaborative perception can happen at different stages of the standard perception network. These stages are similar to those defined by the multi-modal sensor fusion approach \cite{fusion_1}. Each stage of this fusion has its pros and cons; however, in practice, late-stage fusion is more common in industry, and intermediate-stage collaboration is currently more common in academia.

\begin{figure}[!ht]
  \centering
  \includegraphics[width=0.48\textwidth]{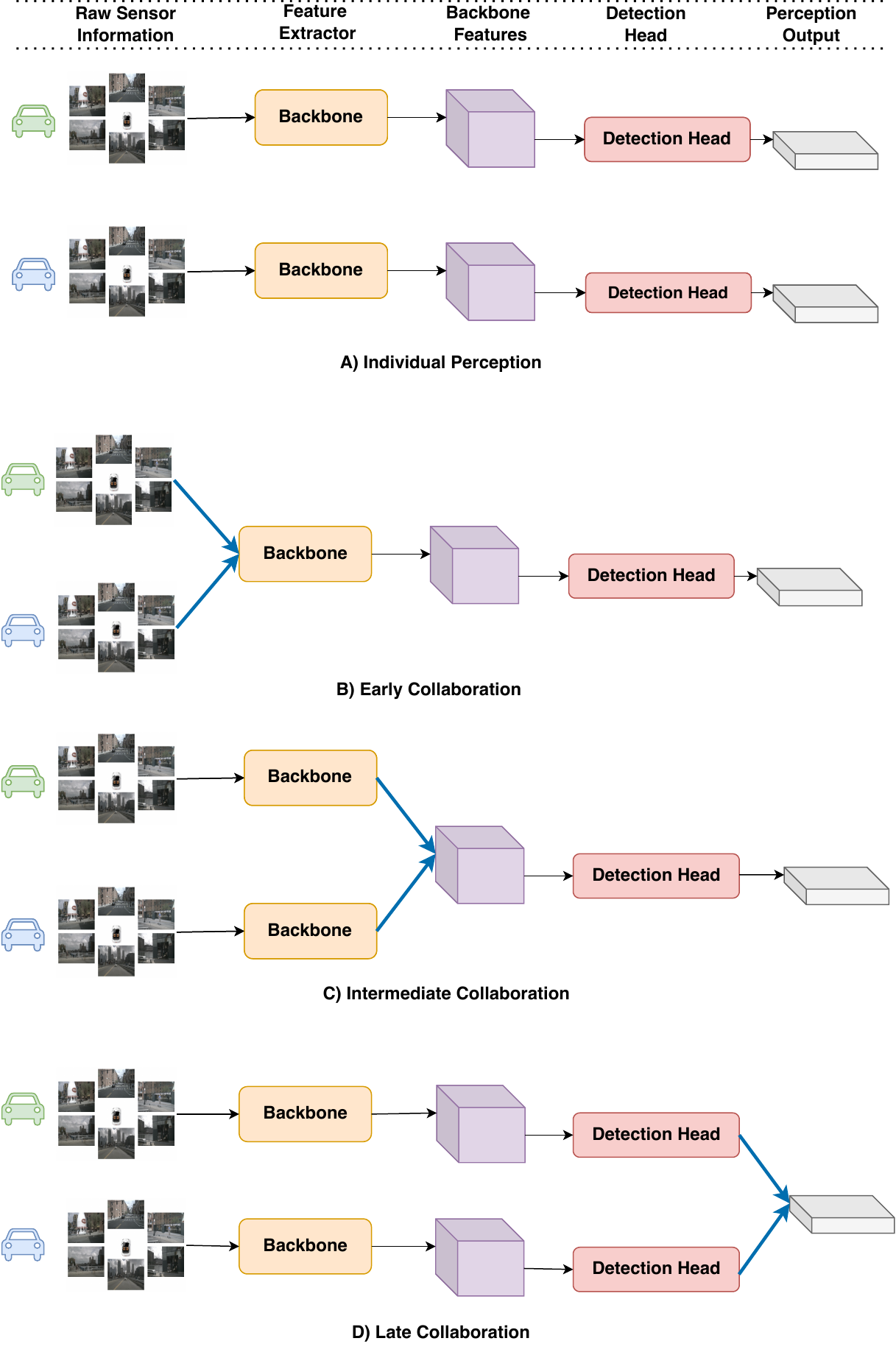}
  \caption{Different stages of collaboration perception, depending on how much information is processed in the perception network before sharing information.}
  \label{fig:fusion_1}
\end{figure}

\textbf{A. Individual Perception}: This is the baseline for our collaboration approach, where there is no collaboration amongst different robots/ vehicles in the robotic fleet. In this case, each of the robots operates solely on their conscience. \\ \\
\textbf{B. Early Collaboration}: This is the simplest approach where each robot in the robotic fleet streams data to every other nearby robot. Then, each ego robot individually processes all the self and others' raw sensor input through its perception network, which includes the backbone and detection head. Despite its simplicity, this approach has a major drawback with the latency in communication and feature computation redundancy at every robot. \\ \\ 
\textbf{C. Intermediate Collaboration}: Here, the goal is to extract meaningful and actionable information by integrating data from different sensors, such as cameras, LiDAR, RADAR, inertial sensors, GPS, and others through the feature extractor, a.k.a; backbone. Then, data-sharing and fusion happen at this actionable information or, in other words, deep learning feature level at every robot. This approach has the best performance currently and is faster than early-stage collaboration. However, it is slower than late-stage collaboration. In practical terms, the bandwidth required to share deep learning features is still quite high, per the current network latency. 
\\ \\
\textbf{D. Late Collaboration}: In this approach, data is shared at the final perception output level, which typically includes 2D/ 3D object bounding boxes or a 2D BEV/ perspective map of semantic segmentation. Here, the fusion process includes data association using Hungarian matching and extrapolation using Kalman filter as shown in \cite{sort}.
\\ \\

\subsection{Why do we need Collaboration Perception?}
\begin{figure}[!ht]
  \centering
  \includegraphics[width=0.48\textwidth]{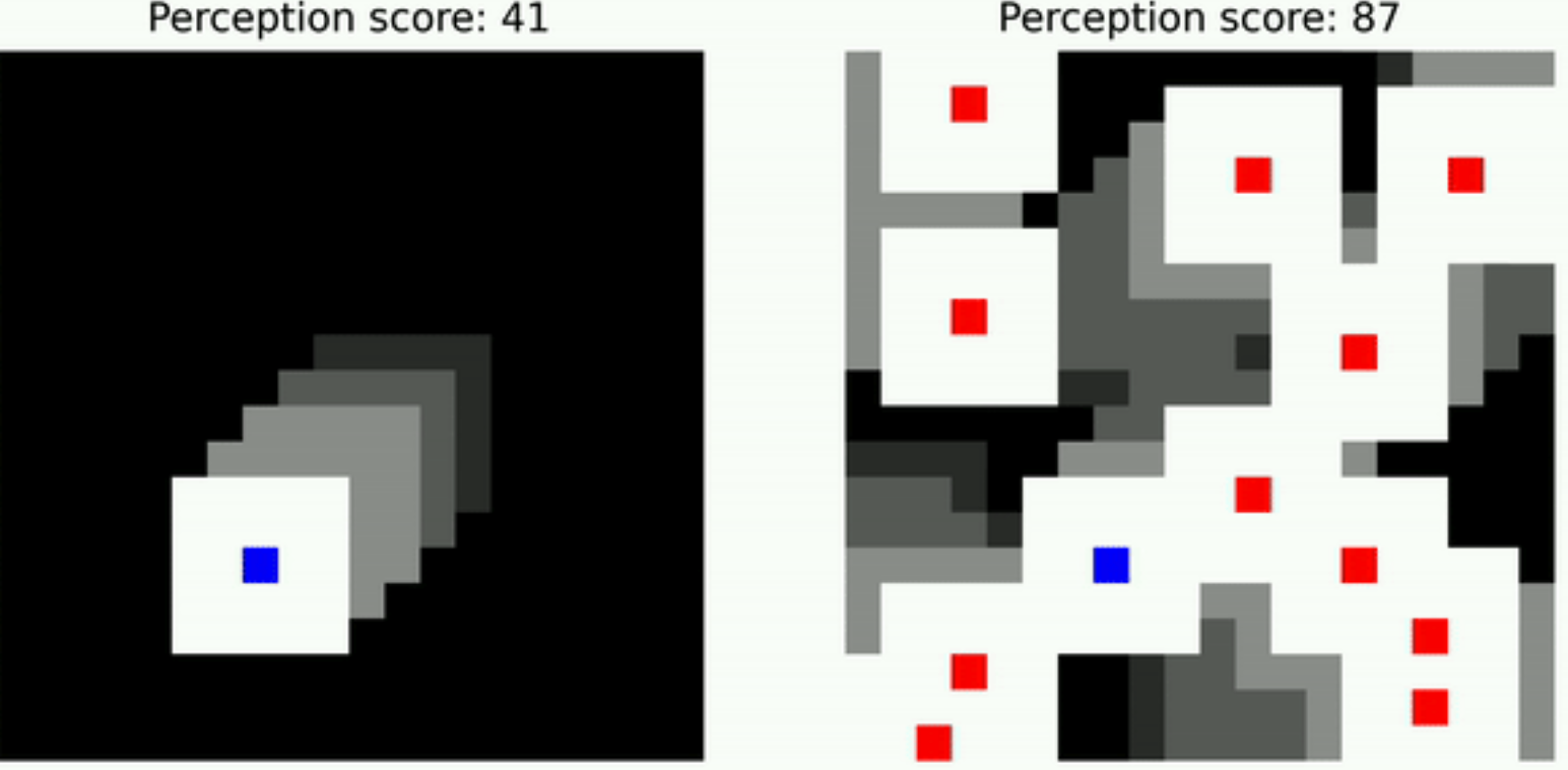}
  \caption{Synpase \cite{synapsebev} visualization. [Left]: Individual perception data. [Right]: Collaborative perception data. Key: Blue: Ego robot; Red: Other robots; Gray-scale: Visibility (Black is completely invisible and white is completely visible).}
  \label{fig:synapsebev}
\end{figure}

In addition to the research papers' performance comparison on benchmark datasets (Simulated and real-world data), we created the most straightforward possible collaboration perception use case and open-sourced our Python library: https://github.com/synapsemobility/synapseBEV. In Fig. \cref{fig:synapsebev}, we see single robot perception data on the left and collaborative perception data on the right using a multi-robot fleet. With this most straightforward demonstration, performance increased by 200\% using collaborative perception of 10+ robotic agents. To simplify this understanding, for the perception score, we just calculated the aggregated sum of the 10*10 grid centered at the ego-robot shown in blue color. Here, we also assume that past perception data is not lost at the instant but is slowly decayed with time. That's why we see a trailing gray shadow behind all the robots. 

\subsection{Problem Driven Collaboration Methods}
We have taken a new approach of dividing the research papers based on the challenging problem they are solving in the space of collaborative perception. Some of the key problems are: 
\begin{itemize}
    \item High communication bandwidth
    \item Agent localization error
    \item Latency/ Communication issues
    \item Model/ Task Discrepancies
    \item Privacy/ Cybersecurity
\end{itemize}

\subsubsection{High Communication Bandwidth}
As we highlighted before, communication bandwidth is one of the major concerns, especially in early and intermediate collaboration. This topic is one of the most discussed in the literature \cite{when2com, where2comm, who2com, keypointsbased}. \cite{who2com} is one of the earliest papers in the chain, where they formulate the problem as one where learned information must be shared across a set of agents in a bandwidth-sensitive manner to optimize for scene understanding tasks such as semantic segmentation. Their approach is built upon network communication protocol. They propose a multi-stage handshake communication mechanism where the neural network can learn to compress relevant information needed for each stage. Specifically, a target agent with degraded sensor data sends a compressed request, the other agents respond with matching scores, and the target agent determines who to connect with (i.e., receive information from). \\ \\
\cite{when2com} is the next paper in the series, which proposes a communication framework by learning both to construct communication groups and decide when to communicate. They demonstrate the generalizability of their framework on two different perception tasks and show that it significantly reduces communication bandwidth while maintaining superior performance. \cite{where2comm} is the last paper in the series. It specifically targets the spatial aspect of the problem. They propose a spatial confidence map, which reflects the spatial heterogeneity of perceptual information. It empowers agents only to share spatially
sparse yet perceptually critical information, contributing to where to communicate. They claim two advantages: (1) it considers pragmatic compression and uses less communication to achieve higher perception performance by focusing on perceptually critical areas, and (2) it can handle varying communication bandwidth by dynamically adjusting spatial areas involved in communication. They evaluated their model on all the latest benchmark datasets \cite{opv2v, dairv2x, v2xsim}. 

\subsubsection{Agent Localization Error}
In many cases, collaboration perception has some errors associated with the agent's GPS locations shared amongst each other. These errors in the shared position and pose cause an inaccurate relative transform
estimation and disrupt the robust mapping of the Ego vehicle. \cite{localization} proposes a distributed object-level cooperative perception system called OptiMatch, in which the detected 3D bounding boxes and
local state information are shared between the connected vehicles. To correct the noisy relative transform, the local measurements of both connected vehicles (bounding boxes) are utilized, and an optimal transport theory-based algorithm is developed to filter out those objects jointly detected by the vehicles along with
their correspondence, constructing an associated co-visible set. \\ \\
On a similar front, \cite{co-allign} targets pose estimation errors due to imperfect localization data of the agents. They propose CoAlign, a hybrid collaboration framework that is robust to the unknown pose errors. The approach relies on agent-object pose graph modeling to enhance pose consistency among collaborating agents. They also adopt a multiscale data fusion strategy to aggregate intermediate features at multiple spatial resolutions. They showed the robustness of their approach by evaluating multiple datasets \cite{opv2v, v2xset, dairv2x}. \cite{keypointsbased} targets localization error with an intermediate collaboration approach.

\subsubsection{Latency/ Communication Issues}
Most of the intermediate papers assume that there is an ideal communication between them despite the high bandwidth required for feature sharing. \cite{lossy} explore the effects of lossy communication (LC) on cooperative perception and propose a novel approach to mitigate these effects. Their approach includes an LC-aware Repair Network (LCRN) and a V2V Attention Module (V2VAM) with intra-vehicle attention and uncertainty-aware inter-vehicle attention. They show their results on OPV2V simulated dataset \cite{opv2v}. \cite{interruption} tackles a similar problem, focussing on interruptions caused by imperfect V2X communication, where cooperation agents can not receive cooperative messages, leading to safety risks. They propose INterruption-aware COoperative Perception (V2X-INCOP), a cooperative perception system robust to communication interruption for V2X communication-aided autonomous driving, which leverages historical cooperation information to recover missing information due to the interruptions and alleviate the impact of the interruption issue. Specifically, they designed a communication-adaptive multi-scale spatial-temporal prediction model to extract multiscale spatial-temporal features based on V2X communication conditions and capture the most significant information to predict the missing information. They also leverage a knowledge distillation framework to give explicit and direct supervision to the prediction model and a curriculum learning strategy to stabilize the model's training. \\ \\
\cite{syncnet} addresses performing robust and reliable time-delay estimation in audio signals in noisy and reverberating environments. Specifically, they propose a semi-causal convolutional neural network of causal and anti-causal layers with a novel correlation-based objective function. The causality in the network ensures non-leakage of representations from future time intervals, and the proposed loss function makes the network generate sequences with high correlation at the actual time delay. \cite{dairv2x} and \cite{v2xfit} also target this challenge with late fusion and intermediate fusion, respectively.

\subsubsection{Model/ Task Discrepancies}
Most of the past research assumes that all the perception systems of the agents follow a common ML model in the back end. This is a fair assumption if the deployment is in a closed environment, like for a multi-robot warehouse; however, for public roads with autonomous vehicles - this assumption is certainly an invalid one. If we use different perception models, we must standardize the confidence scores and data representation from them to the commonly accepted standard. Model-agnostic multi-agent perception framework \cite{modelagnostic} proposed an approach to reduce the negative effect caused by the model discrepancies without sharing the model information. Specifically, they propose a confidence calibrator to eliminate the prediction confidence bias. In their approach, each agent performs such calibration independently on a standard public dataset to protect intellectual property. This paper focuses on early collaboration, which is more intuitive to grasp. However, more research has been done in the intermediate fusion as well \cite{bridging, li2022multirobot}.

\subsubsection{Privacy/ Cybersecurity}
Convolutional neural networks (CNNs), the ML engine behind most common perception approaches, are generally trained on many closed-sourced private data. Authors of \cite{privacy_1} claim that this sensitive data can be leaked through model weights and inference outputs if shared with public resources. They devise an approach by marrying Homographic Encryption (HE) and CNNs to create privacy-preserving neural networks. However, they also mention that this encryption in the CNNs does add significant overheads that grow with the network size. This research has been further extended in \cite{privacy_2, privacy_3, federated}.\\
On the other hand, \cite{adversarial} explores how shared information can be prevented from modifications to execute adversarial attacks on deep learning models that are widely employed in modern systems. They studied the robustness of such systems and focused on exploring adversarial attacks in a novel multi-agent setting where communication is done through sharing learned intermediate representations of neural networks.

\section{Evaluation metrics for vision generative models}
\label{Evaluations and Testing}

\subsection{Evaluation Methods}

We will focus on the most common task of perception in this paper, i.e., 3D object detection. 3D object detectors use multiple criteria to measure the performance of the detectors, viz., precision and recall. However, mean Average Precision (mAP) is the most common evaluation metric. Intersection over Union (IoU) is the ratio of the area of overlap and the area of the union between the predicted box and ground-truth box. An IoU threshold value (generally 0.5) is used to judge if a prediction box matches any particular ground truth box. If IoU is greater than the threshold, then that prediction is treated as a True Positive (TP), else as a False Positive (FP). A ground-truth object which fails to detect with any prediction box, is treated as a False Negative (FN). Precision is the fraction of relevant instances among retrieved instances, while recall is the fraction of retrieved instances.
\begin{equation}
Precision=TP/(TP+FP)
\end{equation}
\begin{equation}
Recall=TP/(TP+FN)
\end{equation}
Based on the above equations, average precision is computed separately for each class. To compare performance between different detectors (mAP) is used. It is a weighted mean based on the number of ground-truths per class. \\ \\
In addition, there are a few dataset-specific metrics viz., KITTI \cite{kitti} introduces Average Orientation Similarity (AOS), which evaluates the quality of orientation estimation of boxes on the ground plane. mAP metric only considers the 3D position of the objects; however, it ignores the effects of both dimension and orientation. In relation to that, nuScenes \cite{nuscenes} introduces TP metrics viz., Average Translation Error (ATE), Average Scale Error (ASE), and Average Orientation Error (AOE). WOD \cite{wod} introduces Average Precision weighted by heading (APH) as its main metric. It takes heading/ orientation information into the account as well. Also, given the depth of confusion for 2D sensors like cameras, WOD introduces Longitudinal Error Tolerant 3D Average Precision(LET-3D-AP), which emphasizes lateral errors more than longitudinal errors in predictions. \\ 

The second most common perception task for robots is semantic segmentation. The evaluation metric commonly used for semantic segmentation is the Intersection over Union (IoU), also known as the Jaccard index. This metric measures the overlap between the predicted segmentation and the ground truth by calculating the intersection of the predicted and ground truth regions divided by their union. The IoU measures how well the predicted segmentation aligns with the actual segmentation, with higher values indicating better performance in accurately capturing the correct regions.

\subsection{Testing Platforms}
There are two main approaches to testing collaborative perception systems. The first is to use pre-collected datasets, which could be real-world or simulated collected datasets. The second is to generate live datasets using a simulator according to the robotic fleet requirement. Generally, robotic fleet requirements are very diverse, based on the deployment site, data-sharing constraints, and number of deployed robots; hence, the second approach is more common in the industry. However, academia tends to rely more on the first approach of benchmark datasets to compare and beat the SOTA network architecture.

\subsubsection{Datasets}
Many open-source datasets use a single vehicle like \cite{nuscenes, wod}. However, industry and academia recently started picking up the more complicated ones with multi-agent collaboration. These datasets are very well appreciated in the industry because syncing the data collected from two independent cars is difficult. Most common datasets are mentioned in Tab. \cref{dataset-benchmark}.

\begin{table*}[t]
\caption{\cite{survey_han}. Benchmark dataset comparison table as of March 2024.}
\label{dataset-benchmark}
\vskip 0.15in
\begin{center}
\begin{small}
\begin{sc}
\begin{tabular}{lcccccccccr}
\toprule
\textbf{Dataset} & \textbf{Conference} & \textbf{Frames} & \textbf{Source} & \textbf{\#agents} & \textbf{Camera} & \textbf{Depth-Camera} & \textbf{LiDAR} \\
\midrule
V2V-Net \cite{v2vnet}  & ECCV'20 & 51k & Simulator & 7 & NA & NA & \checkmark & \\
V2X-SIM \cite{v2xsim} & RAL'21 & 50k & Simulator & 5 & \checkmark & \checkmark & \checkmark  \\
OPV2V \cite{opv2v} & ICRA'22 & 11k & Simulator & 7 & \checkmark & \checkmark & -  \\
DAIR-V2X-C \cite{dairv2x} & CVPR'22 & 39k & Real-data & 2 & \checkmark & \checkmark & -  \\
V2XSet \cite{v2xset} & ECCV'22 & 11k & Simulator & 5 & \checkmark & \checkmark & -  \\
DOLPHINS \cite{dolphins} & ACCV'22 & 42k & Simulator & 3 & \checkmark & \checkmark & -  \\
V2V4Real \cite{v2v4real} & CVPR'23 & 20k & Real-data & 2 & \checkmark & \checkmark & -  \\
V2X-Seq \cite{v2xseq} & CVPR'23 & 15k & Real-data & 2 & \checkmark & \checkmark & -  \\
DeepAccident \cite{deepaccident} \cite{v2xseq} & NA & 15k & Simulator & 5 & \checkmark & \checkmark & -  \\
\bottomrule
\end{tabular}
\end{sc}
\end{small}
\end{center}
\vskip -0.1in
\end{table*}

\begin{table*}[t]
\caption{Cooperative perception for 3D object detection with the Real Data, V2V4REAL \cite{v2v4real}. Two values are shown are the average precision calculated at IoU=0.5/0.7.}
\label{v2v4real}
\vskip 0.15in
\begin{center}
\begin{small}
\begin{sc}
\begin{tabular}{lccccccr}
\toprule
\textbf{Method} & \textbf{Detection Overall} & \textbf{Detection 0-30m} & \textbf{Detection 30-50m} & \textbf{Detection 50-100m} \\
\midrule
No Fusion  & 39.8/22.0 & 69.2/42.6 & 29.3/14.4 & 4.8/1.6  \\
\midrule
Late Fusion & 55.0/26.7 & 73.5/36.8 & 43.7/22.2 & 36.2/17.3  \\
\midrule
Early Fusion  & 59.7/32.1 & 76.1/46.3 & 42.5/20.8 & 47.6/21.1  \\
\midrule
F-Cooper \cite{fcooper} & 60.7/31.8 & 80.8/46.9 & 45.6/23.6 & 32.8/13.4  \\
V2VNet \cite{v2vnet} & 64.5/34.3 & 80.6/51.4 & 52.6/26.6 & 42.6/14.6  \\
AttFuse \cite{opv2v} & 64.7/33.6 & 79.8/44.1 & 53.1/29.3 & 43.6/19.3  \\
V2X-ViT \cite{v2xvit} & 64.9/36.9 & 82.0/55.3 & 51.7/26.6 & 43.2/16.2  \\
CoBEVT \cite{cobevt} & 66.5/36.0 & 82.3/51.1 & 52.1/28.3 & 49.1/19.2  \\
\bottomrule
\end{tabular}
\end{sc}
\end{small}
\end{center}
\vskip -0.1in
\end{table*}

\begin{table*}[t]
\caption{3D object detection comparison with the simulated Data: OPV2V \cite{opv2v}}
\label{opv2v}
\vskip 0.15in
\begin{center}
\begin{small}
\begin{sc}
\begin{tabular}{lcccccccccr}
\toprule
\textbf{Method} & \textbf{Modality} & \textbf{Collaboration Stage} & \textbf{Detection Overall} \\
\midrule
No Fusion  & LiDAR & NA & 60.2   \\
\midrule
Late Fusion & LiDAR & Late & 78.1   \\
\midrule
Early Fusion  & LiDAR & Early & 80.0  \\
\midrule
F-Cooper \cite{fcooper} & LiDAR & Intermediate & 79.0  \\
V2VNet \cite{v2vnet} & LiDAR & Intermediate & 82.2  \\
AttFuse \cite{opv2v} & LiDAR & Intermediate & 81.5  \\
CoBEVT \cite{cobevt} & LiDAR & Intermediate & 83.6  \\
\bottomrule
\end{tabular}
\end{sc}
\end{small}
\end{center}
\vskip -0.1in
\end{table*}

\subsubsection{Simulators}
Since collaborative perception is a relatively newer field, the industry tends to rely on Nvidia Isaac and CARLA as the more common go-to for benchmarking. They are simulators for autonomous mobile robots and autonomous vehicles, respectively. Both of these simulators support multi-robot collaborations. SUMO simulator is another commonly used dataset for autonomous vehicles. Modern simulators are GPU-hungry but can produce well-defined targeted edge cases without concerns about data sync amongst sensors and vehicles. However, results from these simulators must be taken with a grain of salt, as they are not very good at modeling real-world noise and attacks.
\section{Applications in Vision}
\label{applications}

The main applications for collaborative perception lie in mobile robots like autonomous driving, autonomous mobile robots, and drones. However, it also extends to other surveillance computer vision industries with multiple viewpoints. 

\subsection{Autonomous/ Driver-assist Vehicles}
As we have seen in Fig. \cref{fig:car_1}, with the help of collaborative perception, any smart car can see through the physical line of sight boundaries as long as the environment is dense enough with other smart agents/ cars. By communicating data between vehicles, collaborative perception extends the perception capabilities of individual cars, allowing them to detect obstacles, plan paths, and make driving decisions more effectively. This shared perception helps address challenges like occlusion, sparse data, and sensor failures, ultimately enhancing the accuracy of environmental perception and improving overall traffic safety and efficiency.

\subsection{Autonomous Mobile Robots (AMRs)}
One of the most common AMRs is warehouse robots, as shown in Fig. \cref{fig:warehouse_1}, which include autonomous forklifts. Due to on-board computation limitations in these robots they have a fairly small perception range, which can be significantly extended with collaborative perception. Moreover, due to the densely packed nature of any warehouse, which is also shared with humans, there are a lot of blind spots around the corner. Collaborative perception helps reduce these short-range blind spots in robots. The warehouse is one of the most exciting spaces for collaborative perception. All the robots generally work on a common intranet and have a tightly packed robotic fleet in a fairly small geospatial location.

\subsection{Drones}
A collaborative approach enables drones in a swarm to communicate and exchange data to perceive environments collectively, extending their field of view and improving accuracy in detecting obstacles and navigating complex scenarios.

\subsection{Surveillance Cameras}
Smart surveillance cameras system has multi-view cameras linked to ML-based computer vision models. These models are used to query a particular scene or even detect anomalies. All these fleet of cameras linked to the common system can use collaborative perception for spatial awareness of the environment based on their field of view. This setup is already being used in sports analytics.

\section{Challenges \& Future Directions}

\subsection{Challenges}
Collaborative perception in autonomous driving faces several significant challenges that must be addressed to realize its full potential. Some of the major challenges highlighted in the sources include:
\begin{itemize}
    \item \textbf{Data Governance}: Differences in perception models used by different vehicles can lead to inconsistencies in shared sensory data. Challenges arise from discrepancies in sensor types, algorithms, training data, and model parameters among vehicles, impacting the accuracy and reliability of collaborative perception systems. 
    
    \item \textbf{Model Discrepancies}: Differences in perception models used by different vehicles can lead to inconsistencies in shared sensory data. Challenges arise from discrepancies in sensor types, algorithms, training data, and model parameters among vehicles, impacting the accuracy and reliability of collaborative perception systems.
    
    \item \textbf{Communication Capacity and Noise}: Real-time and robust collaborative perception requires addressing communication capacity and noise challenges. Ensuring reliable communication infrastructure and technologies like Vehicle-to-Everything (V2X) communication is essential for effective vehicle collaboration.

    \item \textbf{Privacy and Confidentiality}: Sharing model parameters among autonomous vehicles can raise privacy and confidentiality concerns, especially when vehicles are developed by different manufacturers. Addressing issues related to privacy while sharing sensitive information is crucial for maintaining trust and security in collaborative perception systems. This area has been touched upon in only a handful of papers so far \cite{adversarial, federated}.
    
    \item \textbf {Domain Gap and Confidence Scores}: Disparities in confidence scores among different perception models can impact the performance of collaborative perception systems. Addressing domain gaps in shared sensory data and ensuring consistency in confidence scores are essential for maintaining the effectiveness of collaborative perception.

\end{itemize}

\subsection{Future Directions}
With the rapid development of autonomous robots and vehicles, collaborative perception will play a critical role in revolutionizing and enhancing the capabilities of robotic fleets. However, a large space still exists for exploration in this field. 
\begin{itemize}
    \item \textbf{Foundational Models}: Researchers may leverage the recent popular multi-modal foundation models to enhance the robustness of collaborative perception. Moreover, researchers can try to include the collaboration of perception and prediction models together since they are highly correlated.
    \item \textbf{Fast Database Modelling}: Most current research focuses only on a few robots. However, a real-life robotic fleet will have thousands, if not millions, of robots working together. Researchers must look at ways to store the database and query perception data in O(1) to ensure the shared data is still temporally relevant. This optimizations need to be performed in addition to the current work of communication bandwidth optimization \cite{who2com, when2com, where2comm, keypointsbased}.
    \item \textbf{Data Encryption-Decryption}: There hasn't been a work that encodes vehicle tags in the shared perception data. This kind of encoding is especially required when data is shared between multiple fleets, for example, on the road between automated cars. This data-encoding would help with liability issues and pinpoint and fix any failure case's causes. In addition, such encoding would help validate the data routed toward the ego autonomous vehicles. 
\end{itemize}

\section{Conclusion}
In conclusion, the collaborative perception among multi-robot agents stands as a pivotal advancement in robotics research with far-reaching implications across various domains. Through the synthesis of sensor data, fusion algorithms, and cooperative decision-making mechanisms, these systems exhibit remarkable capabilities in navigating complex environments, enhancing situational awareness, and achieving collective goals efficiently. By leveraging the strengths of individual agents and harnessing the power of collaboration, multi-robot systems can overcome inherent limitations and scale their performance to tackle diverse challenges. However, this field is not without its challenges, including issues related to communication, coordination, and scalability. Addressing these challenges requires continued research efforts in areas such as sensor fusion techniques, communication protocols, and distributed control algorithms. Thus, while significant progress has been made, the journey toward fully realizing the potential of collaborative perception among multi-robot agents is ongoing, promising exciting opportunities for innovation and impact in the realm of robotics and beyond.
{
    \small
    \bibliographystyle{ieeenat_fullname}
    \bibliography{main}

\begin{thebibliography}{37}
\providecommand{\natexlab}[1]{#1}
\providecommand{\url}[1]{\texttt{#1}}
\expandafter\ifx\csname urlstyle\endcsname\relax
  \providecommand{\doi}[1]{doi: #1}\else
  \providecommand{\doi}{doi: \begingroup \urlstyle{rm}\Url}\fi

\bibitem[Apoorv~Singh(2024)]{synapsebev}
Alka~Choudhary Apoorv~Singh.
\newblock Synapsebev.
\newblock \url{https://github.com/synapsemobility/synapseBEV}, 2024.
\newblock Accessed: 2010-09-30.

\bibitem[Bewley et~al.(2016)Bewley, Ge, Ott, Ramos, and Upcroft]{sort}
Alex Bewley, Zongyuan Ge, Lionel Ott, Fabio Ramos, and Ben Upcroft.
\newblock Simple online and realtime tracking.
\newblock In \emph{2016 IEEE International Conference on Image Processing (ICIP)}. IEEE, 2016.

\bibitem[Bi et~al.(2022)Bi, Xiong, Tian, Li, and Liu]{privacy_2}
Renwan Bi, Jinbo Xiong, Youliang Tian, Qi Li, and Ximeng Liu.
\newblock Edge-cooperative privacy-preserving object detection over random point cloud shares for connected autonomous vehicles.
\newblock \emph{IEEE Transactions on Intelligent Transportation Systems}, 23\penalty0 (12):\penalty0 24979--24990, 2022.

\bibitem[Bi et~al.(2023)Bi, Xiong, Tian, Li, and Choo]{privacy_3}
Renwan Bi, Jinbo Xiong, Youliang Tian, Qi Li, and Kim-Kwang~Raymond Choo.
\newblock Achieving lightweight and privacy-preserving object detection for connected autonomous vehicles.
\newblock \emph{IEEE Internet of Things Journal}, 10\penalty0 (3):\penalty0 2314--2329, 2023.

\bibitem[Caesar et~al.(2020)Caesar, Bankiti, Lang, Vora, Liong, Xu, Krishnan, Pan, Baldan, and Beijbom]{nuscenes}
Holger Caesar, Varun Bankiti, Alex~H Lang, Sourabh Vora, Venice~Erin Liong, Qiang Xu, Anush Krishnan, Yu Pan, Giancarlo Baldan, and Oscar Beijbom.
\newblock nuscenes: A multimodal dataset for autonomous driving.
\newblock In \emph{Proceedings of the IEEE/CVF conference on computer vision and pattern recognition}, pages 11621--11631, 2020.

\bibitem[Chellapandi et~al.(2023)Chellapandi, Yuan, Zak, and Wang]{federated}
Vishnu~Pandi Chellapandi, Liangqi Yuan, Stanislaw H~/. Zak, and Ziran Wang.
\newblock A survey of federated learning for connected and automated vehicles, 2023.

\bibitem[Chen(2019)]{fcooper}
Qi Chen.
\newblock F-cooper: Feature based cooperative perception for autonomous vehicle edge computing system using 3d point clouds, 2019.

\bibitem[Geiger et~al.(2012)Geiger, Lenz, and Urtasun]{kitti}
Andreas Geiger, Philip Lenz, and Raquel Urtasun.
\newblock Are we ready for autonomous driving? the kitti vision benchmark suite.
\newblock In \emph{Conference on Computer Vision and Pattern Recognition (CVPR)}, 2012.

\bibitem[Han et~al.(2023)Han, Zhang, Li, Jin, Lang, and Li]{survey_han}
Yushan Han, Hui Zhang, Huifang Li, Yi Jin, Congyan Lang, and Yidong Li.
\newblock Collaborative perception in autonomous driving: Methods, datasets, and challenges.
\newblock \emph{IEEE Intelligent Transportation Systems Magazine}, 2023.

\bibitem[Hu et~al.(2022)Hu, Fang, Lei, Zhong, and Chen]{where2comm}
Yue Hu, Shaoheng Fang, Zixing Lei, Yiqi Zhong, and Siheng Chen.
\newblock Where2comm: Communication-efficient collaborative perception via spatial confidence maps, 2022.

\bibitem[Huang et~al.(2022)Huang, Shi, Li, Li, Huang, and Li]{fusion_1}
Keli Huang, Botian Shi, Xiang Li, Xin Li, Siyuan Huang, and Yikang Li.
\newblock Multi-modal sensor fusion for auto driving perception: A survey, 2022.

\bibitem[Huang et~al.(2023)Huang, Liu, Zhou, Nguyen, Azghadi, Xia, Han, and Sun]{v2xfit}
Tao Huang, Jianan Liu, Xi Zhou, Dinh~C. Nguyen, Mostafa~Rahimi Azghadi, Yuxuan Xia, Qing-Long Han, and Sumei Sun.
\newblock V2x cooperative perception for autonomous driving: Recent advances and challenges, 2023.

\bibitem[Li et~al.(2023)Li, Xu, Liu, Ma, Chi, Ma, and Yu]{lossy}
Jinlong Li, Runsheng Xu, Xinyu Liu, Jin Ma, Zicheng Chi, Jiaqi Ma, and Hongkai Yu.
\newblock Learning for vehicle-to-vehicle cooperative perception under lossy communication.
\newblock \emph{IEEE Transactions on Intelligent Vehicles}, 8\penalty0 (4):\penalty0 2650–2660, 2023.

\bibitem[Li et~al.(2022{\natexlab{a}})Li, Ma, An, Wang, Zhong, Chen, and Feng]{v2xsim}
Yiming Li, Dekun Ma, Ziyan An, Zixun Wang, Yiqi Zhong, Siheng Chen, and Chen Feng.
\newblock V2x-sim: Multi-agent collaborative perception dataset and benchmark for autonomous driving, 2022{\natexlab{a}}.

\bibitem[Li et~al.(2022{\natexlab{b}})Li, Zhang, Ma, Wang, and Feng]{li2022multirobot}
Yiming Li, Juexiao Zhang, Dekun Ma, Yue Wang, and Chen Feng.
\newblock Multi-robot scene completion: Towards task-agnostic collaborative perception.
\newblock In \emph{6th Annual Conference on Robot Learning}, 2022{\natexlab{b}}.

\bibitem[Liu et~al.(2020{\natexlab{a}})Liu, Tian, Glaser, and Kira]{when2com}
Yen-Cheng Liu, Junjiao Tian, Nathaniel Glaser, and Zsolt Kira.
\newblock When2com: Multi-agent perception via communication graph grouping.
\newblock In \emph{Proceedings of the IEEE/CVF Conference on computer vision and pattern recognition}, pages 4106--4115, 2020{\natexlab{a}}.

\bibitem[Liu et~al.(2020{\natexlab{b}})Liu, Tian, Ma, Glaser, Kuo, and Kira]{who2com}
Yen-Cheng Liu, Junjiao Tian, Chih-Yao Ma, Nathan Glaser, Chia-Wen Kuo, and Zsolt Kira.
\newblock Who2com: Collaborative perception via learnable handshake communication, 2020{\natexlab{b}}.

\bibitem[Lu et~al.(2023)Lu, Li, Liu, Dianati, Feng, Chen, and Wang]{co-allign}
Yifan Lu, Quanhao Li, Baoan Liu, Mehrdad Dianati, Chen Feng, Siheng Chen, and Yanfeng Wang.
\newblock Robust collaborative 3d object detection in presence of pose errors, 2023.

\bibitem[Mao et~al.(2023)Mao, Guo, Jia, Sun, Zhou, and Niu]{dolphins}
Ruiqing Mao, Jingyu Guo, Yukuan Jia, Yuxuan Sun, Sheng Zhou, and Zhisheng Niu.
\newblock \emph{DOLPHINS: Dataset for Collaborative Perception Enabled Harmonious and Interconnected Self-driving}, page 495–511.
\newblock Springer Nature Switzerland, 2023.

\bibitem[Raina and Arora(2022)]{syncnet}
Akshay Raina and Vipul Arora.
\newblock Syncnet: Using causal convolutions and correlating objective for time delay estimation in audio signals.
\newblock \emph{arXiv preprint arXiv:2203.14639}, 2022.

\bibitem[Ren et~al.(2024)Ren, Lei, Wang, Dianati, Wang, Chen, and Zhang]{interruption}
Shunli Ren, Zixing Lei, Zi Wang, Mehrdad Dianati, Yafei Wang, Siheng Chen, and Wenjun Zhang.
\newblock Interruption-aware cooperative perception for v2x communication-aided autonomous driving, 2024.

\bibitem[Song et~al.(2023)Song, Wen, Zhang, and Li]{localization}
Zhiying Song, Fuxi Wen, Hailiang Zhang, and Jun Li.
\newblock A cooperative perception system robust to localization errors, 2023.

\bibitem[Sun et~al.(2020)Sun, Kretzschmar, Dotiwalla, Chouard, Patnaik, Tsui, Guo, Zhou, Chai, Caine, Vasudevan, Han, Ngiam, Zhao, Timofeev, Ettinger, Krivokon, Gao, Joshi, Zhao, Cheng, Zhang, Shlens, Chen, and Anguelov]{wod}
Pei Sun, Henrik Kretzschmar, Xerxes Dotiwalla, Aurelien Chouard, Vijaysai Patnaik, Paul Tsui, James Guo, Yin Zhou, Yuning Chai, Benjamin Caine, Vijay Vasudevan, Wei Han, Jiquan Ngiam, Hang Zhao, Aleksei Timofeev, Scott Ettinger, Maxim Krivokon, Amy Gao, Aditya Joshi, Sheng Zhao, Shuyang Cheng, Yu Zhang, Jonathon Shlens, Zhifeng Chen, and Dragomir Anguelov.
\newblock Scalability in perception for autonomous driving: Waymo open dataset, 2020.

\bibitem[Tu et~al.(2021)Tu, Wang, Wang, Manivasagam, Ren, and Urtasun]{adversarial}
James Tu, Tsunhsuan Wang, Jingkang Wang, Sivabalan Manivasagam, Mengye Ren, and Raquel Urtasun.
\newblock Adversarial attacks on multi-agent communication, 2021.

\bibitem[Wang et~al.(2023)Wang, Kim, Ji, Xie, Ge, Chen, Li, and Luo]{deepaccident}
Tianqi Wang, Sukmin Kim, Wenxuan Ji, Enze Xie, Chongjian Ge, Junsong Chen, Zhenguo Li, and Ping Luo.
\newblock Deepaccident: A motion and accident prediction benchmark for v2x autonomous driving, 2023.

\bibitem[Wang et~al.(2020)Wang, Manivasagam, Liang, Yang, Zeng, Tu, and Urtasun]{v2vnet}
Tsun-Hsuan Wang, Sivabalan Manivasagam, Ming Liang, Bin Yang, Wenyuan Zeng, James Tu, and Raquel Urtasun.
\newblock V2vnet: Vehicle-to-vehicle communication for joint perception and prediction, 2020.

\bibitem[Wingarz et~al.(2022)Wingarz, Gomez-Barrero, Busch, and Fischer]{privacy_1}
Tatjana Wingarz, Marta Gomez-Barrero, Christoph Busch, and Mathias Fischer.
\newblock Privacy-preserving convolutional neural networks using homomorphic encryption.
\newblock In \emph{2022 International Workshop on Biometrics and Forensics (IWBF)}, pages 1--6, 2022.

\bibitem[Xu et~al.(2022{\natexlab{a}})Xu, Tu, Xiang, Shao, Zhou, and Ma]{cobevt}
Runsheng Xu, Zhengzhong Tu, Hao Xiang, Wei Shao, Bolei Zhou, and Jiaqi Ma.
\newblock Cobevt: Cooperative bird's eye view semantic segmentation with sparse transformers, 2022{\natexlab{a}}.

\bibitem[Xu et~al.(2022{\natexlab{b}})Xu, Xiang, Tu, Xia, Yang, and Ma]{v2xset}
Runsheng Xu, Hao Xiang, Zhengzhong Tu, Xin Xia, Ming-Hsuan Yang, and Jiaqi Ma.
\newblock V2x-vit: Vehicle-to-everything cooperative perception with vision transformer, 2022{\natexlab{b}}.

\bibitem[Xu et~al.(2022{\natexlab{c}})Xu, Xiang, Tu, Xia, Yang, and Ma]{v2xvit}
Runsheng Xu, Hao Xiang, Zhengzhong Tu, Xin Xia, Ming-Hsuan Yang, and Jiaqi Ma.
\newblock V2x-vit: Vehicle-to-everything cooperative perception with vision transformer, 2022{\natexlab{c}}.

\bibitem[Xu et~al.(2022{\natexlab{d}})Xu, Xiang, Xia, Han, Li, and Ma]{opv2v}
Runsheng Xu, Hao Xiang, Xin Xia, Xu Han, Jinlong Li, and Jiaqi Ma.
\newblock Opv2v: An open benchmark dataset and fusion pipeline for perception with vehicle-to-vehicle communication, 2022{\natexlab{d}}.

\bibitem[Xu et~al.(2023{\natexlab{a}})Xu, Chen, Xiang, Liu, and Ma]{modelagnostic}
Runsheng Xu, Weizhe Chen, Hao Xiang, Lantao Liu, and Jiaqi Ma.
\newblock Model-agnostic multi-agent perception framework, 2023{\natexlab{a}}.

\bibitem[Xu et~al.(2023{\natexlab{b}})Xu, Li, Dong, Yu, and Ma]{bridging}
Runsheng Xu, Jinlong Li, Xiaoyu Dong, Hongkai Yu, and Jiaqi Ma.
\newblock Bridging the domain gap for multi-agent perception, 2023{\natexlab{b}}.

\bibitem[Xu et~al.(2023{\natexlab{c}})Xu, Xia, Li, Li, Zhang, Tu, Meng, Xiang, Dong, Song, Yu, Zhou, and Ma]{v2v4real}
Runsheng Xu, Xin Xia, Jinlong Li, Hanzhao Li, Shuo Zhang, Zhengzhong Tu, Zonglin Meng, Hao Xiang, Xiaoyu Dong, Rui Song, Hongkai Yu, Bolei Zhou, and Jiaqi Ma.
\newblock V2v4real: A real-world large-scale dataset for vehicle-to-vehicle cooperative perception, 2023{\natexlab{c}}.

\bibitem[Yu et~al.(2022)Yu, Luo, Shu, Huo, Yang, Shi, Guo, Li, Hu, Yuan, and Nie]{dairv2x}
Haibao Yu, Yizhen Luo, Mao Shu, Yiyi Huo, Zebang Yang, Yifeng Shi, Zhenglong Guo, Hanyu Li, Xing Hu, Jirui Yuan, and Zaiqing Nie.
\newblock Dair-v2x: A large-scale dataset for vehicle-infrastructure cooperative 3d object detection, 2022.

\bibitem[Yu et~al.(2023)Yu, Yang, Ruan, Yang, Tang, Gao, Hao, Shi, Pan, Sun, Song, Yuan, Luo, and Nie]{v2xseq}
Haibao Yu, Wenxian Yang, Hongzhi Ruan, Zhenwei Yang, Yingjuan Tang, Xu Gao, Xin Hao, Yifeng Shi, Yifeng Pan, Ning Sun, Juan Song, Jirui Yuan, Ping Luo, and Zaiqing Nie.
\newblock V2x-seq: A large-scale sequential dataset for vehicle-infrastructure cooperative perception and forecasting, 2023.

\bibitem[Yuan et~al.(2022)Yuan, Cheng, and Sester]{keypointsbased}
Yunshuang Yuan, Hao Cheng, and Monika Sester.
\newblock Keypoints-based deep feature fusion for cooperative vehicle detection of autonomous driving, 2022.

\end{thebibliography}
}


\end{document}